\documentclass[runningheads]{llncs}
\usepackage{graphicx}

\usepackage{tikz}
\usepackage{comment}
\usepackage{amsmath,amssymb} 
\usepackage{color}

\usepackage[accsupp]{axessibility}  

\begin{document}
\pagestyle{headings}
\mainmatter
\def\ECCVSubNumber{8026}  

\title{Unsupervised Learning of Efficient Geometry-Aware Neural Articulated Representations} 

\titlerunning{ENARF-GAN}
%
\author{Atsuhiro Noguchi\inst{1} \quad
Xiao Sun\inst{2} \quad
Stephen Lin\inst{2} \quad Tatsuya Harada\inst{1, 3}}
\authorrunning{A. Noguchi et al.}
%
\newcommand{\samelineand}{\qquad}
\institute{~\inst{1} The University of Tokyo \samelineand ~\inst{2} Microsoft Research Asia \samelineand ~\inst{3} RIKEN}
\maketitle

\begin{abstract}
We propose an unsupervised method for 3D geometry-aware representation learning of articulated objects, in which no image-pose pairs or foreground masks are used for training.
Though photorealistic images of articulated objects can be rendered with explicit pose control through existing 3D neural representations, these methods require ground truth 3D pose and foreground masks for training, which are expensive to obtain. We obviate this need by learning the representations with GAN training. The generator is trained to produce realistic images of articulated objects from random poses and latent vectors by adversarial training.
To avoid a high computational cost for GAN training, we propose an efficient neural representation for articulated objects based on tri-planes and then present a GAN-based framework for its unsupervised training. Experiments demonstrate the efficiency of our method and show that GAN-based training enables the learning of controllable 3D representations without paired supervision.
    
\end{abstract}
\section{Introduction}
\begin{figure}[t]
    \centering
    \includegraphics[width=1.0\linewidth]{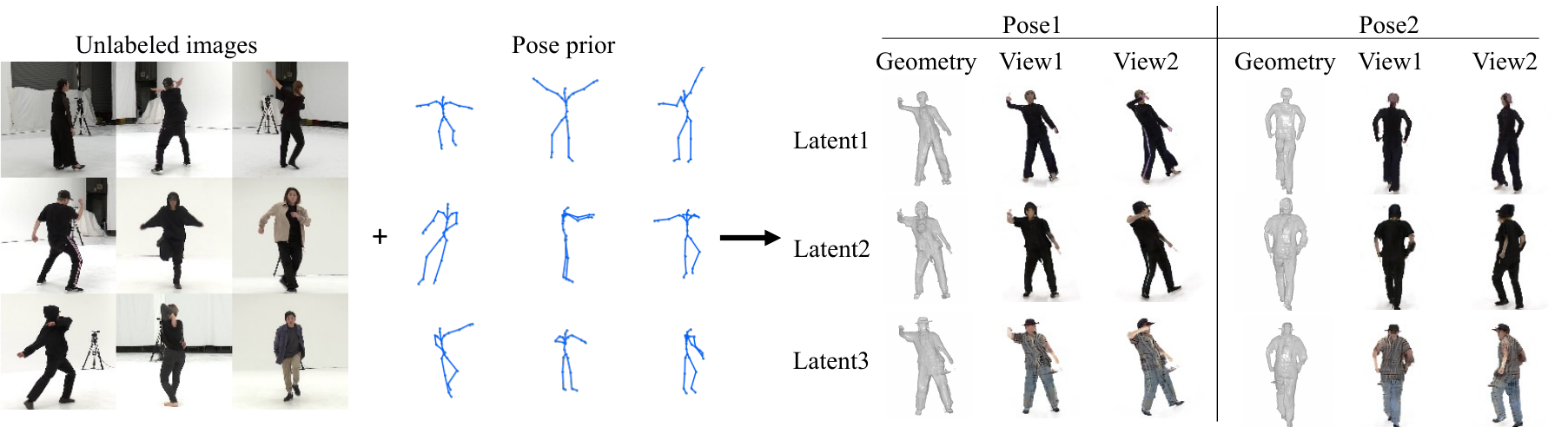}
    \caption{Our ENARF-GAN is a geometry-aware, 3D-consistent image generation model that allows independent control of viewpoint, object pose, and appearance information. It is learned from unlabeled images and a prior distribution on object pose.
    }
    \label{fig:teaser}
\end{figure}

3D models that allow free control over the pose and appearance of articulated objects are essential in various applications, including computer games, media content creation, and augmented/virtual reality. In early work, articulated objects were typically represented by explicit models such as skinned meshes. More recently, the success of learned implicit representations such as neural radiance fields (NeRF)~\cite{mildenhall2020nerf} for rendering static 3D scenes has led to extensions for modeling dynamic scenes and articulated objects. Much of this attention has focused on the photorealistic rendering of humans, from novel viewpoints and with controllable poses, by learning from images and videos.

Existing methods for learning explicitly pose-controllable articulated representations, however, require much supervision, such as 
videos with 3D pose/mesh annotation and a mask for each frame. 
Preparing such data involves tremendous annotation costs; thus, reducing annotation is very important.
In this paper, we propose a novel unsupervised learning framework for 3D pose-aware generative models of articulated objects, which are learned only from unlabeled images of objects sharing the same structure and a pose prior distribution of the objects.

We exploit recent advances in 3D-aware GAN \cite{schwarz2020graf,chan2021pi,niemeyer2021giraffe,gu2022stylenerf,chan2021efficient} for unsupervised learning of the articulated representations. 
They learn 3D-aware image generation models from images without supervision, such as viewpoints or 3D shapes.
The generator is based on NeRF~\cite{mildenhall2020nerf} and is optimized with a GAN objective to generate realistic images from randomly sampled viewpoints and latent vectors from a prior distribution defined before training. 
As a result, the generator learns to generate 3D-consistent images without any supervision.
We employ the idea for articulated objects by defining a pose prior distribution for the target object and optimizing the GAN objective on randomly generated images from random poses and latent variables. It becomes possible to learn a generative model with free control of poses. 
We demonstrate this approach by modeling the pose prior as a skeletal distribution~\cite{noguchi2021neural,su2021nerf}, while noting that other models like meshes~\cite{peng2021neural,peng2021animatable} may bring potential performance benefits.

However, the direct application of existing neural articulated representations to GANs is not computationally practical.
While NeRF can produce high-quality images, its processing is expensive because it requires network inference for every point in space. 
Some methods~\cite{niemeyer2021giraffe,gu2022stylenerf} reduce computational cost by volume rendering at low resolution followed by 2D CNN based upsampling. 
Although this technique achieves high-resolution images with real-time inference speed, it is not geometry-aware (i.e., the surface mesh cannot be extracted).
Recently, a method which we call Efficient NeRF~\cite{chan2021efficient} overcomes the problem.
The method is based on an efficient tri-plane based neural representation and GAN training on it.
Thanks to the computational efficiency, it can produce relatively high-resolution (128 $\times$ 128) images with volumetric rendering.
We extend the tri-plane representation to articulated objects for efficient GAN training.
An overview of the method is visualized in Figure~\ref{fig:teaser}.
The contributions of this work are as follows:
\begin{itemize}
    \item We propose a novel efficient neural representation for articulated objects based on an efficient tri-plane representation. 
    \item We propose an efficient implementation of deformation fields using tri-planes for dynamic scene training, achieving 4 times faster rendering than NARF~\cite{noguchi2021neural} with comparable or better performance.
    \item We propose a novel GAN framework to learn articulated representations without using any 3D pose or mask annotation for each image.
    The controllable 3D representation can be learned from real unlabeled images.
\end{itemize}
\section{Related Work}
\textbf{Articulated 3D Representations.}
The traditional approach for modeling pose-controllable 3D representations of articulated objects is by skinned mesh~\cite{jacobson2014skinning,james2005skinning,lewis2000pose}, where each vertex of the mesh is deformed according to the skeletal pose.
Several parametric skinned mesh models have been developed specifically for humans and animals~\cite{loper2015smpl,hesse2019learning,pavlakos2019expressive,osman2020star,zuffi20173d}. 
For humans, the skinned multi-person linear model (SMPL)~\cite{loper2015smpl} is commonly used.
However, these representations can only handle tight surfaces with no clothing and cannot handle non-rigid or topology-changing objects such as clothing or hair.
Some work alleviates the problem by deforming the mesh surface or using a detailed 3D body scan~\cite{alldieck2018detailed,habermann2021real}.
Recently, implicit 3D shape representations have achieved state-of-the-art performance in pose-conditioned shape reconstruction.
These methods learn neural occupancy/indicator functions~\cite{chen2019learning,mescheder2019occupancy} or signed distance functions~\cite{park2019deepsdf} of articulated objects~\cite{deng2020nasa,bozic2021neural,tiwari2021neural,chen2021snarf}. 
Photorealistic rendering of articulated objects, especially for humans, is also achieved with 3D implicit representations~\cite{peng2021animatable,noguchi2021neural,su2021nerf,alldieck2021imghum,xu2021hnerf,liu2021neural}.
However, all these models require ground truth 3D pose and/or object shape for training.
Very recently, methods have been proposed to reproduce the 3D shape and motion of objects from video data without using 3D shape and pose annotation~\cite{yang2021viser,yang2021banmo,noguchi2021watch}. However, they either do not allow free control of poses or are limited to optimizing for a single object.
An SMPL mesh-based generative model~\cite{grigorev2021stylepeople} and image-to-image translation methods~\cite{esser2018variational,shysheya2019textured,chan2019everybody,liu2020NeuralHumanRendering} can learn pose controllable image synthesis models for humans.
However, the rendering process is completely in 2D and thus is not geometry-aware.

\noindent\textbf{Implicit 3D representations.}
Implicit 3D representations are memory efficient, continuous, and topology free. They have achieved the state-of-the art in learning 3D shape~\cite{chen2019learning,mescheder2019occupancy,park2019deepsdf}, static~\cite{sitzmann2019scene,mildenhall2020nerf,barron2021mip} and dynamic scenes~\cite{pumarola2021d,li2021neural,park2021nerfies}, articulated objects~\cite{peng2021neural,deng2020nasa,bozic2021neural,tiwari2021neural,chen2021snarf,peng2021animatable,noguchi2021neural,su2021nerf,alldieck2021imghum,xu2021hnerf,liu2021neural}, and image synthesis~\cite{schwarz2020graf,chan2021pi}.
Although early works rely on ground truth 3D geometry for training~\cite{chen2019learning,mescheder2019occupancy,park2019deepsdf}, developments in differentiable rendering have enabled learning of networks from only photometric reconstruction losses~\cite{sitzmann2019scene,yariv2020multiview,mildenhall2020nerf}.
In particular, neural radiance fields (NeRF)~\cite{mildenhall2020nerf} applied volumetric rendering on implicit color and density fields, achieving photorealistic novel view synthesis of complex static scenes using multi-view posed images.
Dynamic NeRF~\cite{pumarola2021d,li2021neural,park2021nerfies} extends NeRF to dynamic scenes, but these methods just reenact the motion in the scene and cannot repose objects based on their structure.
Recently, articulated representations based on NeRF have been proposed~\cite{peng2021neural,noguchi2021neural,su2021nerf,peng2021animatable,xu2021hnerf,xu2022surface,liu2021neural}. These methods can render images conditioned on pose configurations. However, all of them require ground truth skeletal poses, SMPL meshes, or foreground masks for training, which makes them unsuitable for in-the-wild images.

Another NeRF improvement is the reduction of computational complexity: NeRF requires forward computation of MLPs to compute color and density for every point in 3D. Thus, the cost of rendering is very high. Fast NeRF algorithms~\cite{garbin2021fastnerf,yu2021plenoctrees,yu_and_fridovichkeil2021plenoxels,mueller2022instant} reduce the computational complexity of neural networks by creating caches or using explicit representations. However, these methods can only be trained on a single static scene.
Very recently, a hybrid explicit and implicit representation was proposed~\cite{chan2021efficient}. In this representation, the feature field is constructed using a memory-efficient explicit representation called a tri-plane, and color and density are decoded using a lightweight MLP. This method can render images at low cost and is well suited for image generation models. 

In this work, we propose an unsupervised learning framework for articulated objects. We extend tri-planes to articulated objects for efficient training.

\noindent\textbf{Generative 3D-aware image synthesis}
Advances in generative adversarial networks (GANs)~\cite{goodfellow2014generative} have made it possible to generate high-resolution, photorealistic images~\cite{karras2019style,karras2020analyzing,karras2021alias}. 
In recent years, many 3D-aware image generation models have been proposed by combining GANs with 3D generators that use meshes~\cite{szabo2019unsupervised}, voxels~\cite{wu2016learning,zhu2018visual,gadelha20173d,henzler2019escaping,nguyen2019hologan,nguyen2020blockgan}, depth~\cite{noguchi2019rgbd}, or implicit representations~\cite{schwarz2020graf,chan2021pi,niemeyer2021giraffe,gu2022stylenerf,chan2021efficient}.
These methods can learn 3D-aware generators without 3D supervision.
Among these, image generation methods using implicit functions, thanks to their continuous and topology-free properties, have been successful in producing 3D-consistent and high-quality images.
However, fully implicit models~\cite{schwarz2020graf,chan2021pi} are computationally expensive, making the training of GANs inefficient.
Therefore, several innovations have been proposed to reduce the rendering cost of generators.
Neural rendering-based methods~\cite{niemeyer2021giraffe,gu2022stylenerf} reduce computation by performing volumetric rendering at low resolution and upsampling the rendered feature images using a 2D CNN. 
Though this enables the generation of high-resolution images at a faster rate, 2D upsampling does not consider 3D consistency and cannot generate detailed 3D geometry.
Very recently, a hybrid of explicit and implicit methods~\cite{chan2021efficient} has been developed for 3d geometry-aware image generation. Instead of using a coordinate-based implicit representation, this method uses tri-planes, which are explicit 3D feature representations, to reduce the number of forward computations of the network and to achieve volumetric rendering at high resolution.

The existing research is specific to scenes of static objects or objects that exist independently of each other, and do not allow free control of the skeletal pose of the generated object. Therefore, we propose a novel GAN framework for articulated objects.

\section{Method}
Recent advances in implicit neural rendering~\cite{su2021nerf,noguchi2021neural} have made it possible to generate 3D-aware pose-controllable images of articulated objects from 
images with accurate 3D pose and foreground mask annotations. However, training such models from only in-the-wild images remains challenging since accurate 3D pose annotations are generally difficult to obtain for them. In the following, we first briefly review Neural Articulated Radiance Field (NARF)~\cite{noguchi2021neural}, then propose an adversarial-based framework, named ENARF-GAN, to efficiently train the NARF model without any paired image-pose and foreground mask annotations.

\subsection{Neural Articulated Radiance Field Revisited}
NARF is an implicit 3D representation for articulated objects. It takes a kinematic 3D pose configuration of an articulated object $o=\{l_k, R_k, {\bf t}_k\}_{k=1:K}$ as input and predicts the color and the density of any 3D location ${\bf x}$, where $l_k$ is the length of the $k^\text{th}$ part, and $R_k$ and ${\bf t}_k$ are its rotation and translation matrices, respectively.
Given the pose configuration $o$, NARF first transforms a global 3D position ${\bf x}$ into several local coordinate systems defined by the rigid parts of the articulated object. Specifically, the transformed local location ${\bf x}_k^l$ for the $k^{\text{th}}$ part is computed as ${\bf x}_k^l = (R^k)^{-1} ({\bf x} - {\bf t}^k) \text{ for } k \in \{1,...,K\}$.

NARF first trains an extra lightweight selector $S$ in the local space to decide which part a global 3D location ${\bf x}$ belongs to. 
Specifically, it outputs the probability $p^k$ of ${\bf x}$ belonging to the $k^{th}$ part.
Then NARF computes color $c$ and density $\sigma$ at the location ${\bf x}$ from a concatenation of local locations masked by the corresponding part probability $p^k$.
\begin{equation}\label{eq: narf_forward}
c, \sigma = G(\text{Cat}(\{\gamma({\bf x}_k^l) * p^k\}_{k=1:K})),
\end{equation}
where $\gamma$ is a positional encoding~\cite{mildenhall2020nerf}, $\text{Cat}$ is the concatenation operation, and $G$ is an MLP network.
The RGB color ${\bf C}$ and foreground mask value ${\bf M}$ for each pixel are generated by volumetric rendering~\cite{mildenhall2020nerf}. The network is trained with a reconstruction loss between the generated and ground truth color $\hat{\bf C}$ and mask $\hat{\bf M}$,
\begin{equation}\label{eq:supervised_loss}
    \mathcal L_{\text{supervised}} = \sum_{r\in \mathcal R}\left(||{\bf C} - \hat {\bf C}||_2^2 + ||{\bf M} - \hat {\bf M}||_2^2\right),
\end{equation}
where $\mathcal R$ is the set of rays in each batch.
Please refer to the original NARF paper~\cite{noguchi2021neural} for more details.

\subsection{Unsupervised Learning by Adversarial Training}
\begin{figure}[t]
\centering
\includegraphics[width=1.0\linewidth]{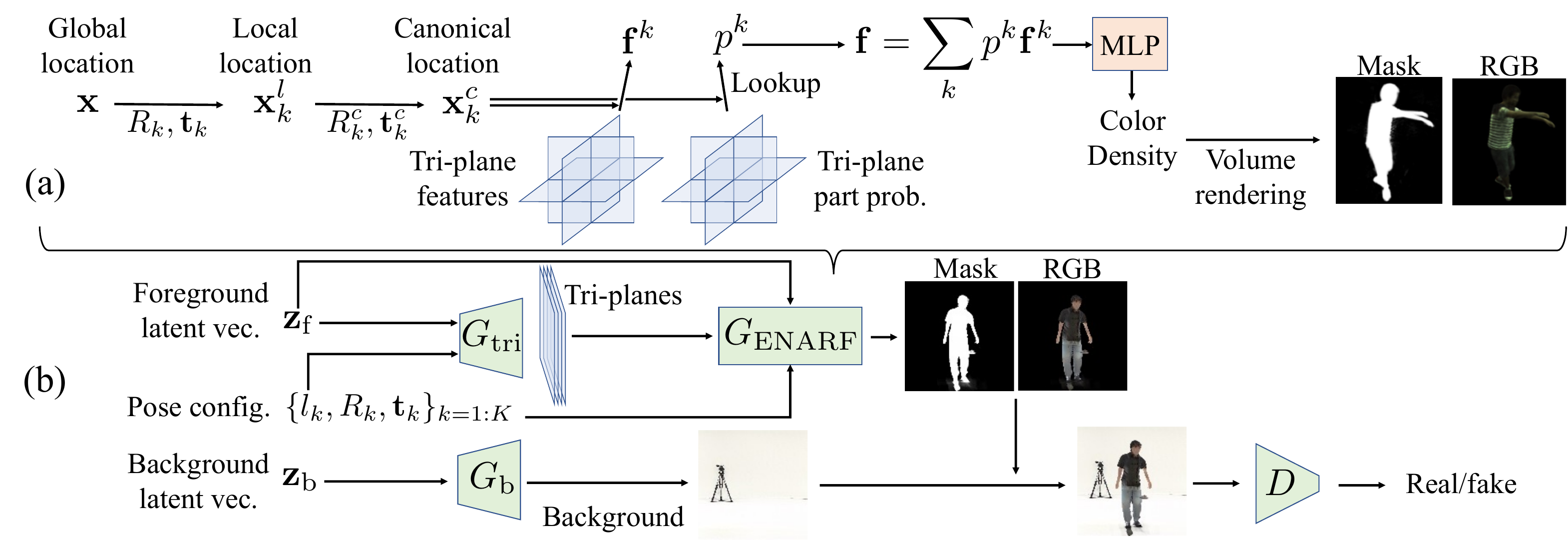}
\caption{Overview of (a) Efficient NARF (ENARF) and (b) GAN training.}
\label{fig:pipeline}
\end{figure}

In this work, we propose a method for efficient and unsupervised training of the NARF model from unposed image collections.
Without loss of generality, we consider humans as the articulated objects here.
To this end, we first define a human pose distribution $\mathcal{O}$. For one training iteration, our NARF based generator $G$ takes a latent vector ${\bf z}$ and a sampled 
human pose instance $o$ from $\mathcal{O}$ as input and predicts a synthesized image ${\bf C}$. Following standard adversarial training of GANs, a discriminator $\mathcal{D}$ is used to distinguish the synthesized image ${\bf C}$ from real ones $\tilde{\bf C}$. Formally, the training objectives of the generator $\mathcal L_{\text{adv}}^G$ and discriminator $\mathcal L_{\text{adv}}^D$ are defined as follows, 
\begin{equation}\label{eq:adv_loss}
    \mathcal L_{\text{adv}}^G = - \mathbb E\left[\log(D(G({\bf z}, o)))\right], \mathcal L_{\text{adv}}^D = -\mathbb E\left[\log(D(\tilde {\bf C})) + \log(1 - D(G({\bf z}, o)))\right].
\end{equation}
An overview of this method is illustrated in Figure~\ref{fig:pipeline} (b).

However, this training would be computationally expensive. The rendering cost of NARF is heavy because computation is performed for many 3D locations in the viewed space. 
Even though in supervised training, the time and memory cost of computing the reconstruction loss could be reduced by evaluating it over just a small proportion of the pixels~\cite{noguchi2021neural}, the adversarial loss in Equation~\ref{eq:adv_loss} requires the generation of the full image for evaluation. As a result, the amount of computation becomes impractical.

In the following, we propose a series of changes in feature computation and the selector to address this issue. Note that these changes not only enable the GAN training
but also greatly improve the efficiency of the original NARF.

\subsection{Efficiency Improvements on NARF}
Recently, Chan et al.~\cite{chan2021efficient} proposed a hybrid explicit-implicit 3D-aware network that uses a memory-efficient tri-plane representation to explicitly store features on axis-aligned planes. With this representation, the efficiency of feature extraction for a 3D location is greatly improved. Instead of forwarding all the sampled 3D points through the network, the intermediate features of arbitrary 3D points can be obtained via simple lookups on the tri-planes.
The tri-plane representation can be more efficiently generated with Convolutional Neural Networks (CNNs) instead of MLPs. The intermediate features of 3D points are then transformed into the color and density using a lightweight decoder MLP.
The decoder significantly increases the non-linearity between features at different positions, thus greatly enhancing the expressiveness of the model.

Here, we adapt the tri-plane representation to NARF for more efficient training.
Similar to~\cite{chan2021efficient}, we first divide the original NARF network $G$ into an intermediate feature generator (the first linear layer of $G$) $W$ and a decoder network $G_{\text{dec}}$. Then, Equation~\ref{eq: narf_forward} is rewritten as follows.
\begin{equation}
\label{eq.split}
    c, \sigma = G_\text{dec}({\bf f}) \text{ , where } {\bf f} = W(\text{Cat}(\{\gamma({\bf x}_k^l) * p^k | k \in \{1,...,K\}\})),
\end{equation}
where ${\bf f}$ is an intermediate feature vector of input 3D location ${\bf x}$.
However,
re-implementing the feature generator $W$ to produce the tri-plane representation is not straightforward because its input, a weighted concatenation of ${\bf x}_k^l$, does not form a valid location in a specific 3D space. 
We make two important changes to address this issue.
First, we decompose $W$ into $K$ sub-matrices $\{W_k\}$, one for each part, where each one takes the corresponding local position ${\bf x}_k^l$ as input and outputs an intermediate feature for the $k^\text{th}$ part.
Then, the intermediate feature in Equation~\ref{eq.split} can be equivalently rewritten as follows.
\begin{equation}\label{eq:rewrite}
    {\bf f} = \sum_{k=1}^K p^k * {\bf f}^k \text{ , where }{\bf f}^k = W_k(\gamma({\bf x}_k^l)),
\end{equation}
where ${\bf f}^k$ is a feature generated in the local coordinate system of the $k^\text{th}$ part. Now, $W_k(\gamma({\bf x}_k^l))$ can be directly re-implemented by 
tri-planes $F$.
However, the computational complexity of this implementation is still proportional to $K$.
In order to train a single tri-plane for all parts, the second change is to further transform the local coordinates ${\bf x}_k^l$ into a 
canonical space defined by a canonical pose $o^c$, similar to Animatable NeRF~\cite{peng2021animatable}.
\begin{equation}\label{eq:canonical}
  {\bf x}_k^c = R_k^c {\bf x}_k^l + {\bf t}_k^c = R_k^c (R^k)^{-1} ({\bf x} - {\bf t}^k) + {\bf t}_k^c,
\end{equation}
where 
$R_k^c$ and ${\bf t}_k^c$ are the rotation and translation matrices of the canonical pose. 
Intuitively, ${\bf x}_k^c$ is the corresponding point location of ${\bf x}$ transformed into the canonical space 
when ${\bf x}$ is considered to belong to the $k^\text{th}$ part. 
Finally, the tri-plane feature $F$ is learned in the canonical space. The feature extraction for location ${\bf x}$ is achieved by
retrieving the 32-dimensional feature vector ${\bf f}^k$ on the tri-plane $F$ at ${\bf x}_k^c$
for all parts, then taking a weighted sum of those features as in Equation~\ref{eq:rewrite},
\begin{equation}
{\bf f} = 
\sum_{k=1}^K p^k * {\bf f}^k \text{ , where }{\bf f}^k 
= \!\!\!\!\!\!\!\sum_{ij \in \{xy, yz, xz\}}\!\!\!\!\!\!\! F_{ij}({\bf x}_k^c)
.
\end{equation}
$F_{**}({\bf a})$ is a retrieved feature vector from each axis-aligned plane at location ${\bf a}$.

We estimate the RGB color $c$ and density $\sigma$ from ${\bf f}$ using a lightweight decoder network $G_\text{dec}$ consisting of three FC layers with a hidden dimension of 64 and output dimension of 4. 
We apply volume rendering~\cite{mildenhall2020nerf} on the color and density to output an RGB image ${\bf C}$ and a foreground mask ${\bf M}$.

Although we efficiently parameterize the intermediate features, 
the probability $p^k$ needs to be computed for every 3D point and every part.
In the original NARF, though lightweight MLPs are used to estimate the probabilities, they are still computationally infeasible.

Therefore, we propose an efficient selector network using tri-planes.
Since $p^k$ is used to mask out features of irrelevant parts, this probability can be a rough approximation of the shape of each part. Thus the tri-plane representation is expressive enough to model the probability.
We use $K$ separate 1-channel tri-planes to represent $P^k$, the part probability projected to each axis-aligned plane. We retrieve the three probability values $(p_{xy}^k, p_{xz}^k, p_{yz}^k) = (P^k_{xy}({\bf x}_k^c), P^k_{xz}({\bf x}_k^c), P^k_{yz}({\bf x}_k^c))$ of the $k^\text{th}$ part by querying the 3D location in the canonical space ${\bf x}_k^c$. 
The probability $p^k$ that ${\bf x}_k^c$ belongs to the $k^\text{th}$ part is approximated as $p^k = p_{xy}^k p_{xz}^k p_{yz}^k$.

In this way, the features $F$ and part probabilities $P$ are modeled efficiently with a single tri-plane representation. The tri-plane is represented by a $(32 + K) \times 3$ channel image. The first 96 channels represent the tri-plane features in the canonical space.
The remaining $3K$ channels represent the tri-plane probability maps
for each of the $K$ parts.
We call this approach Efficient NARF, or ENARF.

\subsection{GAN}
To condition the generator on latent vectors, we utilize a StyleGAN2~\cite{karras2020analyzing} based generator to produce tri-plane features.
We condition each layer of the proposed ENARF by the latent vector with a modulated convolution~\cite{karras2020analyzing}.
Since the proposed tri-plane based generator can only represent the foreground object, we use an additional StyleGAN2 based generator for the background. 

We randomly sample latent vectors for our tri-plane generator and background generator: ${\bf z} = ({\bf z}_\text{tri}, {\bf z}_\text{ENARF}, {\bf z}_b) \sim \mathcal N (0, I)$,
where ${\bf z}_\text{tri}$, ${\bf z}_\text{ENARF}$, and ${\bf z}_b$ are latent vectors for the tri-plane generator, ENARF, and background generator, respectively.
The tri-plane generator $G_\text{tri}$ generates tri-plane feature $F$ and part probability $P$ from randomly sampled ${\bf z}_\text{tri}$ and bone length $\{l_k\}_{k=1:K}$. 
$G_\text{tri}$ takes $l_k$ as inputs to account for the diversity of bone lengths.
\begin{equation}
    F, P = G_\text{tri}({\bf z}_\text{tri}, \{l_k\}_{k=1:K})
\end{equation}
The ENARF based foreground generator $G_\text{ENARF}$ generates the foreground RGB image ${\bf C}_f$ and mask ${\bf M}_f$ from the generated tri-planes, and the background generator $G_\text{b}$ generates background RGB image ${\bf C}_b$. 
\begin{equation}
{\bf C}_f, {\bf M}_f = G_\text{ENARF}({\bf z}_\text{ENARF}, \{l_k, R_k, {\bf t}_k\}_{k=1:K}), {\bf C}_b = G_\text{b}({\bf z}_b)
\end{equation}
The final output RGB image is ${\bf C} = {\bf C}_f + {\bf C}_b * (1 - {\bf M}_f)$, which is a composite of ${\bf C}_f$ and ${\bf C}_b$. 
To handle the diversity of bone lengths, we replace Equation~\ref{eq:canonical} with one normalized by the length of the bone: 
${\bf x}_k^c = \frac{l_k^c}{l_k} R_k^c {\bf x}_k + {\bf t}_k^c$,
where $l^c_k$ is the bone length of the $k^\text{th}$ part in the canonical space.

We optimize these generator networks with GAN training. We use a bone loss in addition to an adversarial loss on images, R1 regularization on the discriminator~\cite{mescheder2018training}, and L2 regularization on the tri-planes.
The bone loss ensures that an object is generated in the foreground. 
Based on the input pose of the object, a skeletal image $B$ is created, where pixels with skeletons are 1 and others are 0, and the generated mask $M$ at pixels with skeletons is made close to 1:
$\mathcal L_{\text{bone}} = \frac{\sum_{r\in \mathcal R}(1 - M)^2B}{\sum_{r\in \mathcal R}B}$.
Additional details are provided in the supplement. 
The final loss is the linear combination of these losses.
\subsection{Dynamic Scene Overfitting}
Since ENARF is an improved version of NARF, we can directly use it for single dynamic scene overfitting.
For training, we use the ground truth 3D pose and foreground mask of each frame and optimize the reconstruction loss in Equation~\ref{eq:supervised_loss}.

If the object shape is strictly determined by the poses that comprise the kinematic motion, we can use the same tri-plane features for the entire sequence and directly optimize them.
However, real-world objects have time or pose-dependent non-rigid deformation such as clothing and facial expression change in a single sequence.
Therefore, the tri-plane features should change depending on time and pose.
We use a technique based on deformation fields~\cite{pumarola2021d,park2021nerfies} proposed in time-dependent NeRF, also known as Dynamic NeRF.
Deformation field based methods learn a mapping network from observation space to canonical space and learn the NeRF in the canonical frame.
Since learning the deformation field with an MLP is expensive, we also approximate it with tri-planes.
We approximate the deformation in 3D space by independent 2D deformations in each tri-plane.
First, a StyleGAN2~\cite{karras2020analyzing} generator takes positionally encoded time $t$ and a rotation matrix of each part $R_k$ and generates 6-channel images representing the relative 2D deformation from the canonical space of each tri-plane feature. 
We deform the tri-plane feature based on the generated deformation. Please refer to the supplement for more details.
We use constant tri-plane probabilities $P$ for all frames since the object shares the same coarse part shape throughout the entire sequence.
The remaining networks are the same. We refer to this method as D-ENARF.

\section{Experiments}
Our experimental results are presented in two parts. First, in Section~\ref{sec.exp.supervised}, we compare the proposed Efficient NARF (ENARF) with the state-of-the-art methods \cite{noguchi2021neural,peng2021animatable} in terms of both efficiency and effectiveness, and we conduct ablation studies on the deformation modeling and the design choices for the selector.
Second, in Section~\ref{sec.exp.gan}, we present our results of using adversarial training on ENARF, namely, ENARF-GAN, and compare it with baselines. Then, we discuss the effectiveness of the pose prior and the generalization ability of ENARF-GAN.

\subsection{Training on a Dynamic Scene}
\label{sec.exp.supervised}
Following the training setting in Animatable NeRF~\cite{peng2021animatable}, we train our ENARF model on synchronized multi-view videos of a single moving articulated object. The ZJU mocap dataset~\cite{peng2021neural} consisting of three subjects (313, 315, 386) is used for training.
We use the same pre-processed data provided by the official implementation of Animatable NeRF.
All images are resized to $512\times512$. 
We use 4 views and the first 80\% of the frames for training, and the remaining views or frames for testing. In this setting, the ground truth camera and articulated object poses, as well as the ground truth foreground mask, are given for each frame. More implementation details can be found in the supplement.

First, we compare our method with the state-of-the-art supervised methods NARF~\cite{noguchi2021neural} and Animatable NeRF~\cite{peng2021animatable}.  Our comparison with Neural Actor~\cite{liu2021neural} is provided in the supplement.
Note that our method and NARF~\cite{noguchi2021neural} take ground truth kinematic pose parameters (joint angles and bone lengths) as inputs, while Animatable NeRF needs the ground truth SMPL mesh parameters.
In addition, Animatable NeRF requires additional training on novel poses to render novel-pose images, which is not necessary for our model.

Table~\ref{tab:dso_quant} shows the quantitative results.
To compare the efficiency between models, we examine the GPU memory,
FLOPS, and the running time used to render an entire image of resolution $512 \times 512$ on a single A100 GPU as evaluation metrics.
To compare the quality of synthesized images under novel view and novel pose settings, 
PSNR,
SSIM~\cite{wang2004image}, and 
LPIPS~\cite{zhang2018unreasonable} are used as evaluation metrics.
Table~\ref{tab:dso_quant} shows that the proposed Efficient NARF achieves competitive or even better performance compared to existing methods with far fewer FLOPS and 4.6 times the speed of the original NARF.
Although the runtime of ENARF is a bit slower than Animatable NeRF (0.05s), its performance is 
superior under both novel view and novel pose settings. In addition, it does not need extra training on novel poses.
Our dynamic model D-ENARF further improves the performance of ENARF with little increased overhead in inference time, and outperforms the state-of-the-arts Animatable NeRF and NARF by a large margin.
\begin{table}[t]
    \caption{Quantitative comparison on dynamic scenes.}
    \label{tab:dso_quant}
    \centering
    \scalebox{0.8}[0.8]{
    \begin{tabular}{l|ccc|ccc|ccc}
      \hline
        & \multicolumn{3}{|c|}{Cost} & \multicolumn{3}{|c|}{Novel view} & \multicolumn{3}{|c}{Novel pose} \\
        & \#Memory& \#FLOPS & Time(s)& PSNR$\uparrow$ & SSIM$\uparrow$ & LPIPS$\downarrow$ &PSNR$\uparrow$ & SSIM$\uparrow$ & LPIPS$\downarrow$ \\ \hline \hline
Animatable NeRF~\cite{peng2021animatable}&-&-&{\bf 0.42}&28.28&0.9484&0.05818&29.09&0.9507&0.05706\\
NARF~\cite{noguchi2021neural}.    &283.9GB&15.9T&2.17&30.62&0.9625&0.05228&29.51&{\bf 0.959}&\underline{0.05208}\\ \hline
ENARF                             &\underline{27.0GB}&\underline{71.7G}&0.47&31.94&0.9655&0.04792&29.66&0.953&0.05702\\
D-ENARF                           &27.6GB&354G&0.49&{\bf 32.93}&{\bf 0.9713}&{\bf 0.03718}&{\bf 30.06}&0.9396&{\bf 0.05205}\\
ENARF w/o selector                    &{\bf 23.0GB}&{\bf 70.2G}&\underline{0.43}&29.16&0.9493&0.07234&27.9&0.9377&0.08316\\
ENARF w/ MLP selector                 &83.2GB&337G&1.13&\underline{32.27}&\underline{0.9684}&\underline{0.04633}&\underline{29.74}&\underline{0.9573}&0.05228\\
      \hline
    \end{tabular}}
  \end{table}
\begin{figure}[t]
    \centering
    \includegraphics[width=1.0\linewidth]{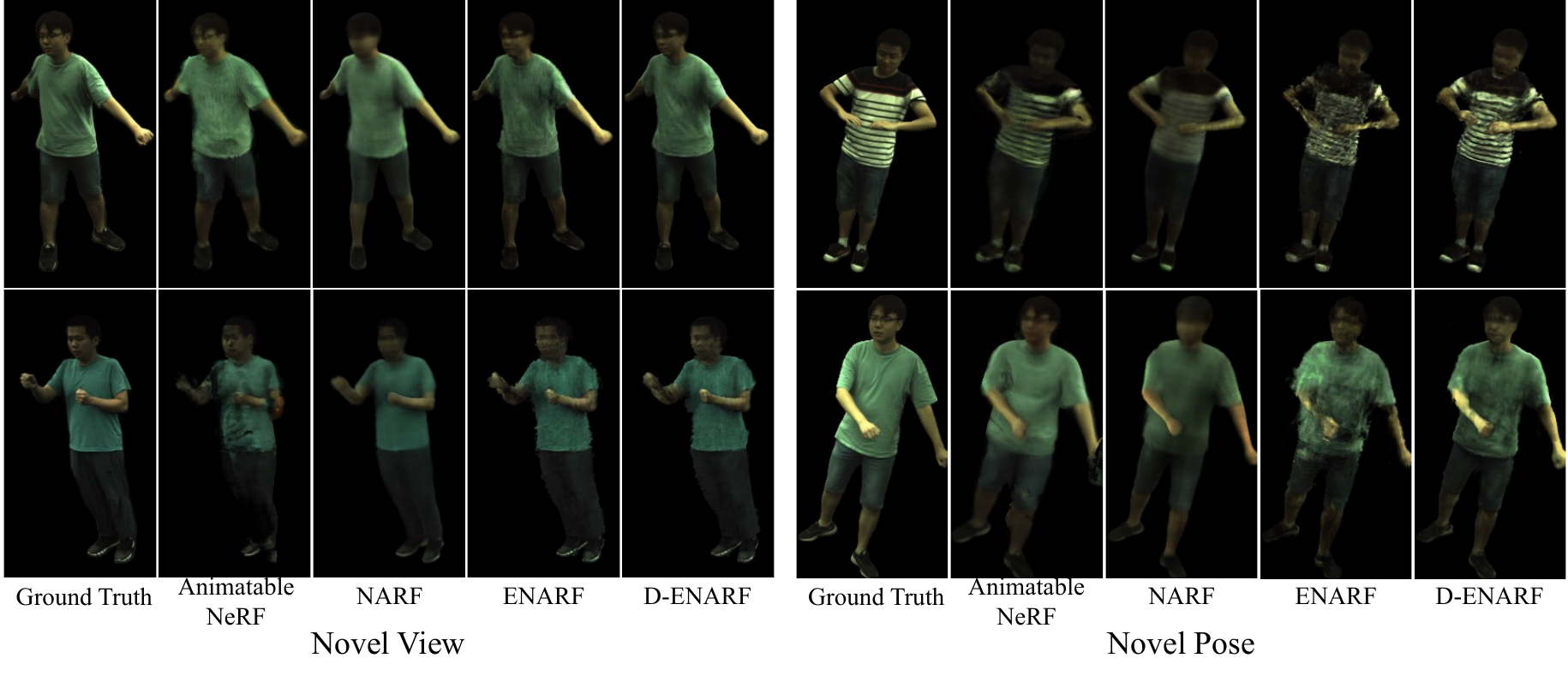}
    \caption{Qualitative comparison of novel view and pose synthesis.}
\label{fig:novel_view_pose_comp}
\end{figure}
Qualitative results for novel view and pose synthesis are shown in Figure~\ref{fig:novel_view_pose_comp}.
ENARF produces much sharper images than NARF due to the more efficient explicit–implicit tri-plane representation.
D-ENARF, which utilizes deformation fields, further improves the rendering quality.
In summary, the proposed D-ENARF method achieves better performance in both
image quality and computational efficiency.

\noindent\textbf{Ablation Study}
To evaluate the effectiveness of the tri-plane based selector, we compare our method against models using an MLP selector or without a selector.
A quantitative comparison is provided in Table~\ref{tab:dso_quant}, and a qualitative comparison is provided in the supplement.
Although an MLP based selector improves the metrics a bit, it 
results in a significant increase in testing time.
In contrast, the model without a selector is unable to learn clean/sharp part shapes and textures, because a 3D location will inappropriately be affected by all parts without a selector.
These results indicate that our tri-plane based selector is efficient and effective.

\subsection{Unsupervised Learning with GAN}
\label{sec.exp.gan}
In this section, we train the proposed efficient NARF using GAN objectives without any image-pose pairs or mask annotations.

\noindent\textbf{Comparison with Baselines}
Since this is the first work to learn an articulated representation without image-pose pairs or mesh shape priors, no exact competitor exists.
We thus compare our method against two baselines. 
The first is a \textit{supervised} baseline called ENARF-VAE, inspired by the original NARF~\cite{noguchi2021neural}. Here, a ResNet50~\cite{he2016deep} based encoder estimates the latent vectors ${\bf z}$ from images, and the efficient NARF based decoder decodes the original images from estimated latent vectors ${\bf z}$ and ground truth pose configurations $o$. 
These networks are trained with the reconstruction loss defined in Equation~\ref{eq:supervised_loss} and the KL divergence loss on the latent vector ${\bf z}$. Following~\cite{noguchi2021neural}, ENARF-VAE is trained with images with a black background.
The second model is called StyleNARF, which is a combination of the original NARF and the state-of-the-art high-resolution 3D-aware image generation model called StyleNeRF~\cite{gu2022stylenerf}.
To reduce the computational cost, the original NARF first generates low-resolution features using volumetric rendering. 
Subsequently, a 2D CNN-based network upsamples them into final images.
Additional details are provided in the supplement.
Please note that ENARF-VAE is a \textit{supervised} method and cannot handle the background, and StyleNARF loses 3D consistency and thus cannot generate high-resolution geometry.

We use the SURREAL dataset~\cite{varol2017learning} for comparison. It is a synthetic human image dataset with a resolution of $128 \times 128$.
Dataset details are given in the supplement.
For the pose prior, we use the ground truth pose distribution of the training dataset, where we randomly sample poses from the entire dataset.
Please note that we do not use image-pose pairs for these unsupervised methods.

\begin{table}[t]
    \caption{Quantitative comparison on generative models. * indicates that the methods are modified from the cited papers.}
    \label{tab:gan_quant}
    \centering
    \scalebox{0.8}[0.8]{
    \begin{tabular}{l|cccc}
      \hline
        & FID$\downarrow$ & FG-FID$\downarrow$ & Depth$\downarrow$ & PCKh@0.5 $\uparrow$ \\ \hline \hline
ENARF-VAE~\cite{noguchi2021neural}*.       &-   &63.0&{\bf 3.2} &{\bf 0.984}\\ 
StyleNARF~\cite{gu2022stylenerf}*.   &{\bf 20.8}&-   &16.5&0.924\\  \hline
ENARF-GAN                             &\underline{22.6}&{\bf 21.3}&\underline{8.8} &\underline{0.947}\\  \hline
ENARF-GAN CMU pri.                    &24.2&\underline{25.6}&12.8&0.915\\
ENARF-GAN rand. pri.                  &25.9&37.0&13.8&0.887\\
\qquad w/ trunc. $\psi=0.4$          &26.5&24.1&8.0 &0.884\\ 
      \hline
    \end{tabular}}
  \end{table}
  
Quantitative results are shown in Table~\ref{tab:gan_quant}.
We measure image quality with the Fr\'echet Inception Distance (FID)~\cite{heusel2017gans}.
To better evaluate the quality of foreground, we use an extra metric called FG-FID that replaces the background with black color, using the generated or ground truth mask.
We measure depth plausibility by comparing the real and generated depth map.
Although there is no ground truth depth for the generated images, the depth generated from a pose would have a similar depth to the real depth that arises from the same pose.
We compare the L2 norm between the inverse depth generated from poses sampled from the dataset and the real inverse depth of them.
Finally, we measure the correspondence between pose and appearance 
following the contemporary work named GNARF~\cite{bergman2022generative}. We apply an off-the-shelf 2D human keypoint estimator~\cite{mmpose2020} to both generated and real images with the same poses and compute the Percentage of Correct Keypoints (PCK) between them, which is commonly used for evaluating 2D pose estimators.
We report the averaged PCKh@0.5 metric~\cite{andriluka20142d} for all keypoints. Details are provided in the supplement.
Qualitative results are shown in Figure~\ref{fig:gan_comp}.
Not surprisingly, ENARF-VAE produces the most plausible depth/geometry and learns the most accurate pose conditioning among the three since it uses image-pose pairs for supervised training. However, compared to styleNARF, its FID is worse and the images lack photorealism.
styleNARF achieves the best FID among the three, thanks to the effective CNN renderer. 
However, it cannot explicitly render the foreground only or generate accurate geometry of the generated images.
In contrast, our method performs volumetric rendering at the output resolution, and the generated geometry perfectly matches the generated foreground image.

\begin{figure}[t]
    \centering
    \includegraphics[width=1.0\linewidth]{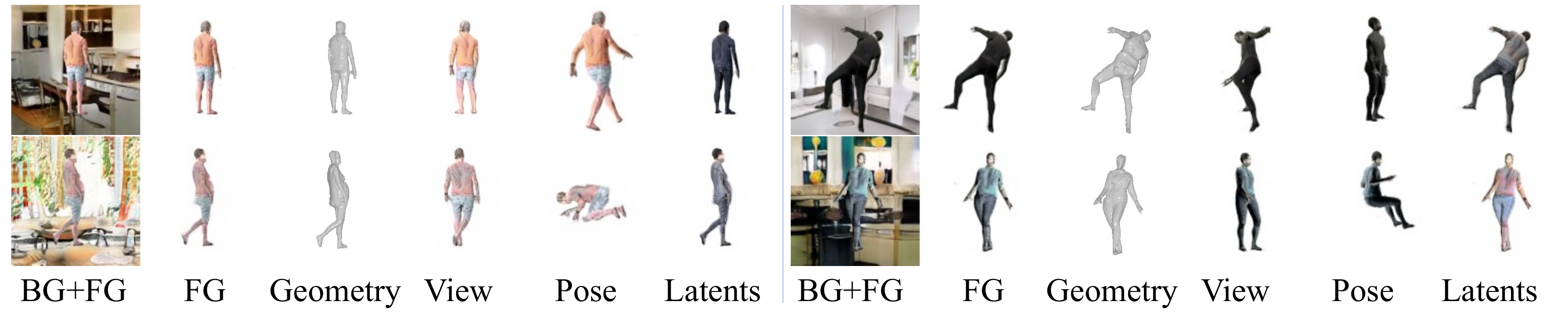}
    \caption{Learned geometry and disentangled representations on the SURREAL dataset by ENARF-GAN. For each of the generated results, the leftmost three columns show the generated images with background, foreground, and corresponding geometry.
The rightmost three images show the results of changing only the viewpoint, object pose, and latent variables for the same results, respectively.
    }
\label{fig:gan_results}
\end{figure}
\begin{figure}[t]
    \centering
    \includegraphics[width=0.9\linewidth]{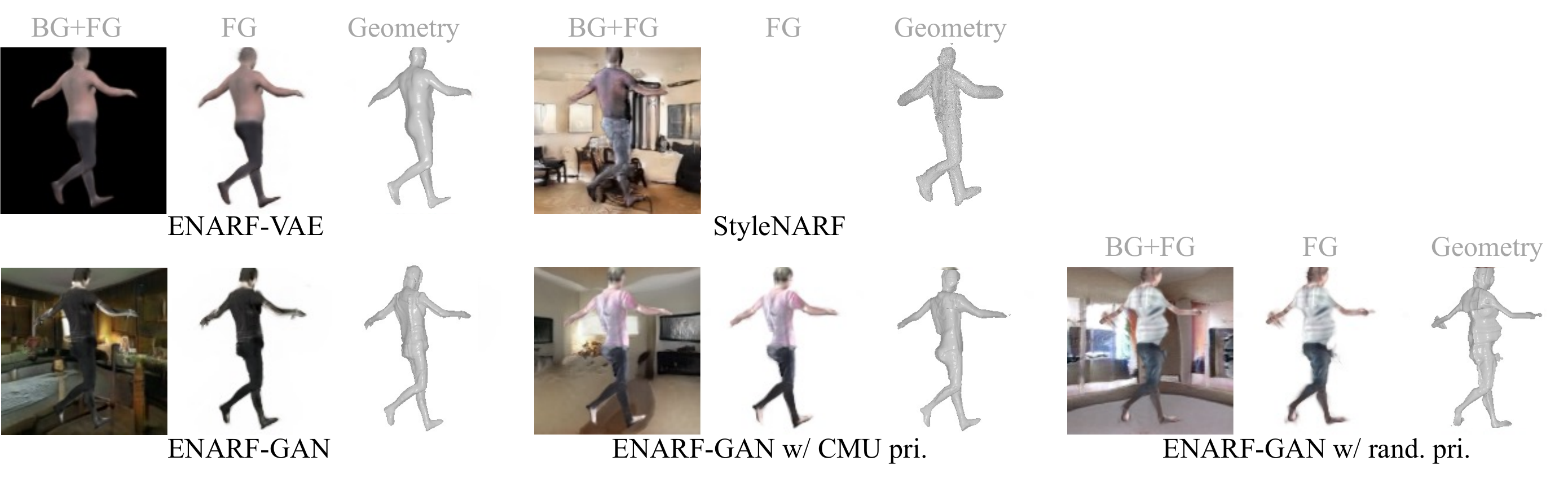}
    \caption{Qualitative comparison on generative models.}
\label{fig:gan_comp}
\end{figure}

\noindent\textbf{Using Different Pose Distribution}
Obtaining a ground truth pose distribution of the training images is not feasible for in-the-wild images or new categories.
Thus, we train our model with a pose distribution different from the training images.
Here, we consider two pose prior distributions.
The first uses poses from CMU Panoptic~\cite{joo2015panoptic} as a prior, which we call the CMU prior.
During training, we randomly sample poses from the entire dataset.
In addition, to show that our method works without collecting actual human motion capture data, we also create a much simpler pose prior.
We fit a multi-variate Gaussian distribution on each joint angle of the CMU Panoptic dataset, and we use randomly sampled poses from the distribution for training. Each Gaussian distribution only defines the rotation angle of each part, which can be easily constructed for novel objects. We call this the random prior.

Quantitative and qualitative results are shown in Table~\ref{tab:gan_quant} and Figure~\ref{fig:gan_comp}.
We can confirm that even when using the CMU prior, our model learns pose-controllable 3D representations with just a slight sacrifice in image quality.
When using the random prior, the plausibility of the generated images and the quality of the generated geometry are worse.
This may be because the distribution of the random prior is so far from the distribution of poses in the dataset that the learned space of latent vectors too often falls outside the distribution of the actual data. Therefore, we used the truncation trick~\cite{karras2019style} to restrict the diversity of the latent space, and the results are shown in the bottom row of Table~\ref{tab:gan_quant}.
By using the truncation trick, even with a simple prior, we can eliminate latent variables outside the distribution and improve the quality of the generated images and geometry.
Further experimental results on truncation are given in the supplement.

\begin{figure}[t]
    \centering
    \includegraphics[width=1.0\linewidth]{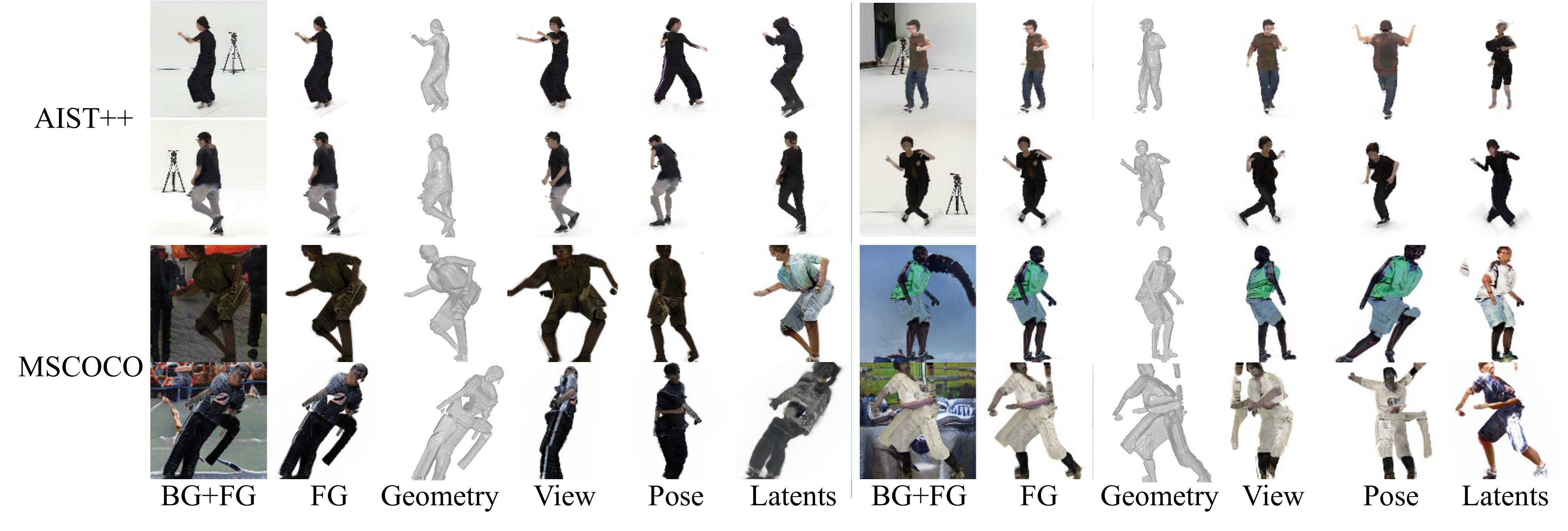}
    \caption{Qualitative results on AIST++ and MSCOCO.}
\label{fig:aist_mscoco}
\end{figure}

\noindent\textbf{Additional Results on Real Images}
To show the generalization ability of the proposed framework, we train our model on two real image datasets, namely AIST++~\cite{li2021ai} and MSCOCO~\cite{lin2014microsoft}.
AIST++ is a dataset of dancing persons with relatively simple backgrounds. We use the ground truth pose distribution for training.
MSCOCO is a large scale in-the-wild image dataset. 
We choose images capturing roughly the whole human body and crop them around the persons. 
Since 3D pose annotations are not available for MSCOCO, we use poses in CMU Panoptic as the pose prior.
Note that we do not use any image-pose or mask supervisions for training.
Qualitative results are shown in Figure~\ref{fig:aist_mscoco}.
Experimental results with AIST++, which has a simple background, show that it is possible to generate detailed geometry and images with independent control of viewpoint, pose, and appearance.
For MSCOCO, two successful and two unsuccessful results are shown in Figure~\ref{fig:aist_mscoco}. MSCOCO is a very challenging dataset because of the complex background, the lack of clear separation between foreground objects and background, and the many occlusions. Although our model does not always produce plausible results, it is possible to generate geometry and control each element independently. As an initial attempt, the results are promising.

\section{Conclusion}
In this work, we propose a novel unsupervised learning framework for 3D geometry-aware articulated representations.
We showed that our framework is able to learn representations with controllable viewpoint and pose. 
We first propose a computationally efficient neural 3D representation for articulated objects by adapting the tri-plane representation to NARF, then show it can be trained with GAN objectives without using ground truth image-pose pairs or mask supervision.
However, the resolution and the quality of the generated images are still limited compared to recent NeRF-based GAN methods; meanwhile, we assume that a prior distribution of the object's pose is available, which may not be easily obtained for other object categories.
Future work includes incorporating the neural rendering techniques proposed in 3D-aware GANs to generate photorealistic high-quality images while preserving the 3D consistency and estimating the pose prior distribution directly from the training data.

\paragraph{\textbf{Acknowledgements}.}
This work was supported by D-CORE Grant from Microsoft Research Asia and partially supported by JST AIP Acceleration Research JPMJCR20U3, Moonshot R\&D Grant Number JPMJPS2011, CREST Grant Number JPMJCR2015, JSPS KAKENHI Grant Number JP19H01115, and JP20H05556 and Basic Research Grant (Super AI) of Institute for AI and Beyond of the University of Tokyo. We would like to thank Haruo Fujiwara, Lin Gu, Yuki Kawana, and the authors of \cite{bergman2022generative} for helpful discussions.

%
%
\bibliographystyle{splncs04}
\bibliography{egbib}

\clearpage

\appendix

\begin{figure}
    \centering
    \vspace{-8mm}
    \includegraphics[width=0.83\linewidth]{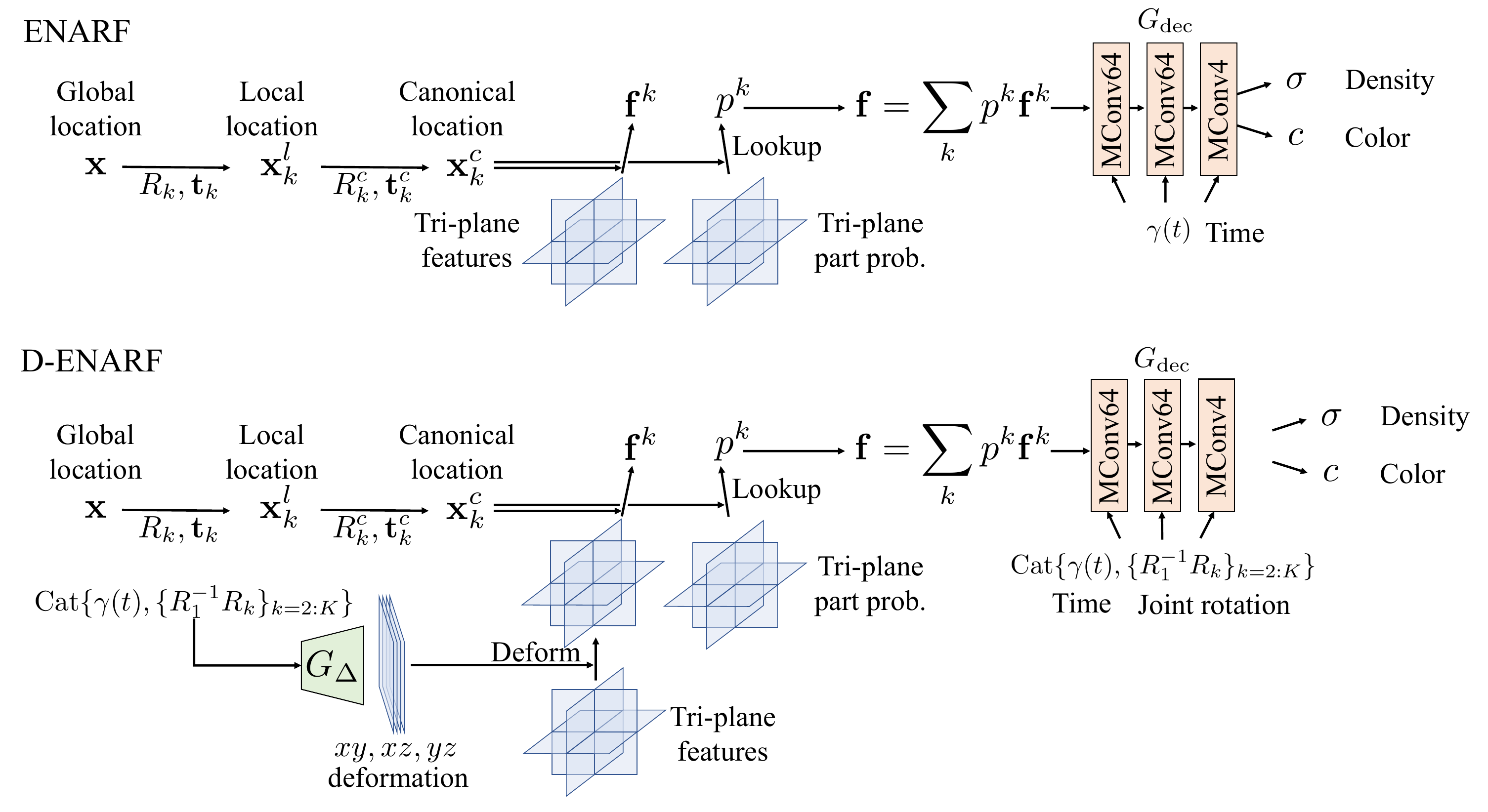}
    \caption{Network details of ENARF and D-ENARF. MConv denotes Modulated Convolution~\cite{karras2020analyzing}.}
    \label{fig:network_details}
    \vspace{-8mm}
  \end{figure}
  
  \section{ENARF Implementation details}
  
  \subsection{Model Details}
  Figure~\ref{fig:network_details} illustrates the learning pipeline of our ENAFR and D-ENAFR models.
  The decoder network $G_\text{dec}$ is implemented with a modulated convolution as in~\cite{karras2020analyzing}. For the ENARF model,  $G_\text{dec}$ is conditioned on the time input $t$ with positional encoding $\gamma(*)$ to handle time-dependent deformations. Following~\cite{mildenhall2020nerf}, we use 10 frequencies in the positional encoding. For the D-ENARF model, rotation matrices are additionally used as input to handle pose-dependent deformations. To reduce the input dimension, we input the relative rotation matrices for each part to the root part $R_1$.
  To efficiently learn the deformation field in our D-ENARF model, additional tri-plane deformation $\Delta$ are learned with a StyleGAN2~\cite{karras2020analyzing} based generator $G_\Delta$,
  \begin{equation}
    \Delta = G_\Delta(\text{Cat}\{\gamma(t), \{R_1^{-1} R_k\}_{k=2:K}\}).
  \end{equation}
Each channel represents the relative transformation from the canonical frame in pixels for each tri-planes. The learned deformation field is used to deform the canonical features to handle the non-rigid deformation for each time and pose. The deformed tri-plane feature $F'$ is formulated as,

\begin{align}\label{eq:deformation}
  F'_{xy}({\bf x}) &= F_{xy}([{\bf x}_x + \Delta_{xy}({\bf x})_x,{\bf x}_y + \Delta_{xy}({\bf x})_y, {\bf x}_z])\\ \notag
  F'_{yz}({\bf x}) &= F_{yz}([{\bf x}_x, {\bf x}_y + \Delta_{yz}({\bf x})_y,{\bf x}_z + \Delta_{yz}({\bf x})_z])\\ \notag
  F'_{xz}({\bf x}) &= F_{xz}([{\bf x}_x + \Delta_{xz}({\bf x})_x, {\bf x}_y, {\bf x}_z + \Delta_{xz}({\bf x})_z]),
\end{align}
where ${\bf x} = [{\bf x}_x, {\bf x}_y, {\bf x}_z]$. Intuitively, $({\bf x}_x, {\bf x}_y)$ is the 2D projection of ${\bf x}$ on the $F_{xy}$ plane, and
$\Delta_{xy}({\bf x})$ produces the 2D translation vector for $({\bf x}_x, {\bf x}_y)$ on the $F_{xy}$ plane.
However, this operation requires two look-up tri-planes learned from $\Delta$ and $F$ for every 3D location ${\bf x}$, which is expensive. Therefore, we make an approximation of this by first computing the value of the deformed tri-plane $F'$ at every pixel grid location using Equation~\ref{eq:deformation}, then sampling features from $F'$ using Equation~7 in the main paper. The increased computational cost is thus proportional to the resolution of $F'$, which is much smaller than the number of 3D points ${\bf x}$.

We used the coarse-to-fine sampling strategy as in~\cite{mildenhall2020nerf} to sample the rendering points on camera rays. Instead of using two separate models to predict points at coarse and fine stages respectively~\cite{mildenhall2020nerf}, we use a single model to predict sampled points at both stages. Specifically, for each ray, 48 and 64 points are sampled at the coarse and fine stages, respectively.
  
  \subsection{Efficient Implementation}
  For efficiency, we introduce a weak shape prior for the part occupancy probability $p^k$ in Section~3.3 of the main paper. Specifically, if a 3D location ${\bf x}_k^c$ in the canonical space is outside of a cube with one side of $2a$ located at the center of the part ${\bf p}_k^c$,
  we set $p^k$ to 0, namely, $p^k \leftarrow 0 \text{ if } \max(|{\bf x}_k^c - {\bf p}_k^c|) > a$. We set $a$ to $\frac{1}{3}$ meter for all parts. This weak shape prior is used for all NARF-based methods.
  Since we do not have to compute the intermediate feature ${\bf f}_k$ (in Equation~7 in the main paper) for the points with part probability $p_k=0$, the overall computational cost for feature generation is significantly reduced.
  We implement this efficiently by (1) gathering the valid ($p_k>0$) canonical positions ${\bf x}_k^c$ and compute intermediate features ${\bf f}_k$ for them, (2) multiplying by $p^k$, and (3) summing up the feature for each part $p_k * {\bf f}_k$ with a scatter\_add operation.
  
  \subsection{Training Details}
  We use the Adam~\cite{kingma2014adam} optimizer with an equalized learning rate~\cite{karras2017progressive} of 0.001. The learning rate decay rate is set to 0.99995. The ray batch sizes are set to 4096 for ENARF and 512 for NARF. The ENARF model is trained for 100,000 iterations and the NARF model is trained for 200,000 iterations with a batch size 16. The training takes 15 hours on a single A100 GPU for ENARF and 24 hours for NARF.
  
  \section{Ablation Study on Selector (Section 4.1)}
\begin{figure}[t]
    \centering
    \includegraphics[width=0.9\linewidth]{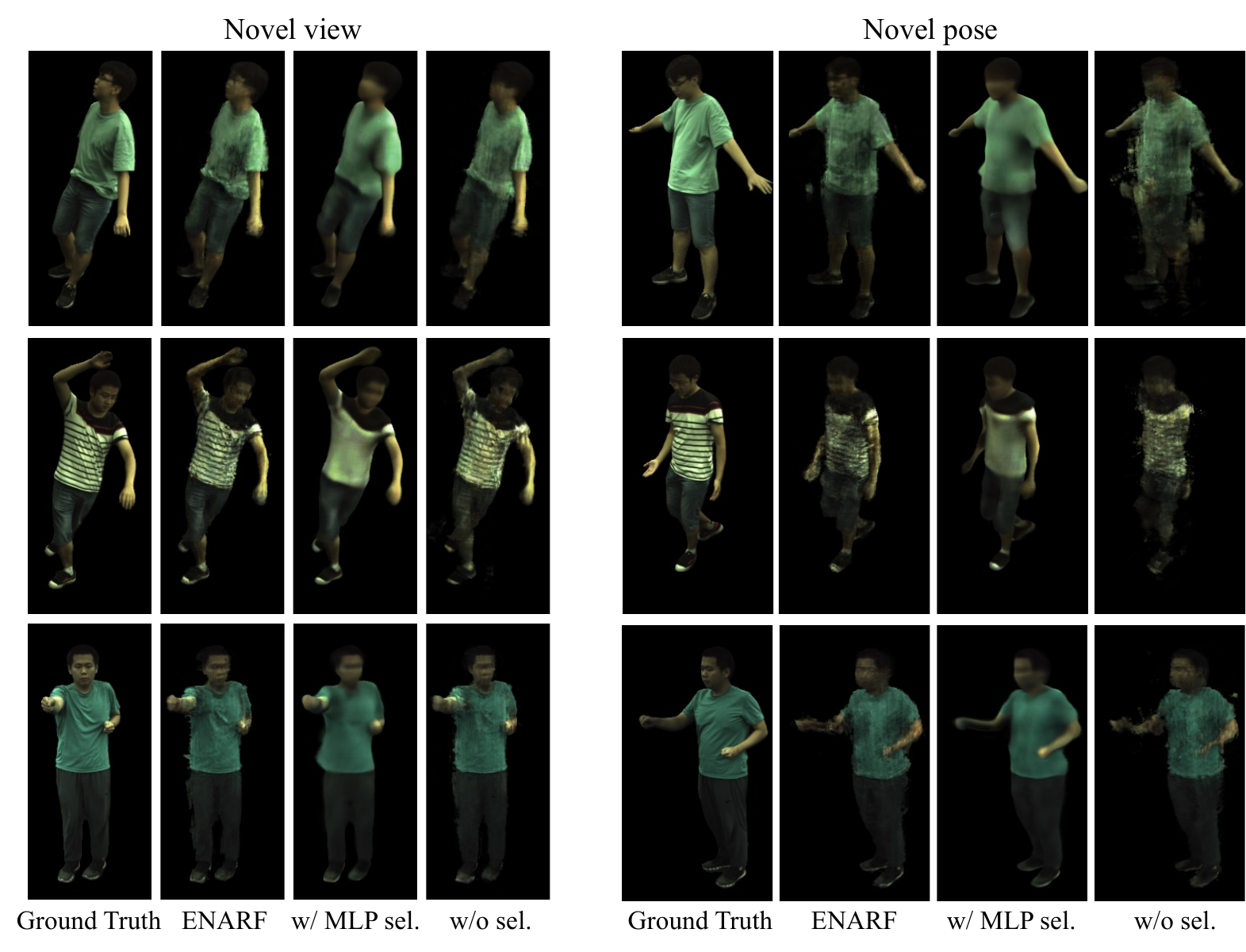}
    \caption{Ablation study on selector.}
  \label{fig:selector_ablation}
  \end{figure}

  To show the effectiveness of the tri-plane based selector, we compared our model with a model without the selector and a model with an MLP based selector.
  In the model without a selector, we simply set $p^k=\frac{1}{K}$.
  In the MLP based selector, we used a two-layer MLP with a hidden layer dimension of 10 for each part, as in NARF~\cite{noguchi2021neural}.
  The quantitative and qualitative comparisons are provided in Table~1 in the main paper and Figure~\ref{fig:selector_ablation}, respectively.
  The model without a selector cannot generate clear and sharp images because the feature of any location to compute the density and color is evenly contributed by all parts.
  The MLP based selector helps learn independent parts. However, the generated images look blurry compared to ours. Moreover, it requires much more GPU memory and FLOPS for training and testing.
  In summary, the proposed tri-plane based selector is superior in terms of both effectiveness and efficiency.
  
  \section{Ablation Study on View Dependency}
\begin{figure}[t]
    \centering
    \includegraphics[width=0.9\linewidth]{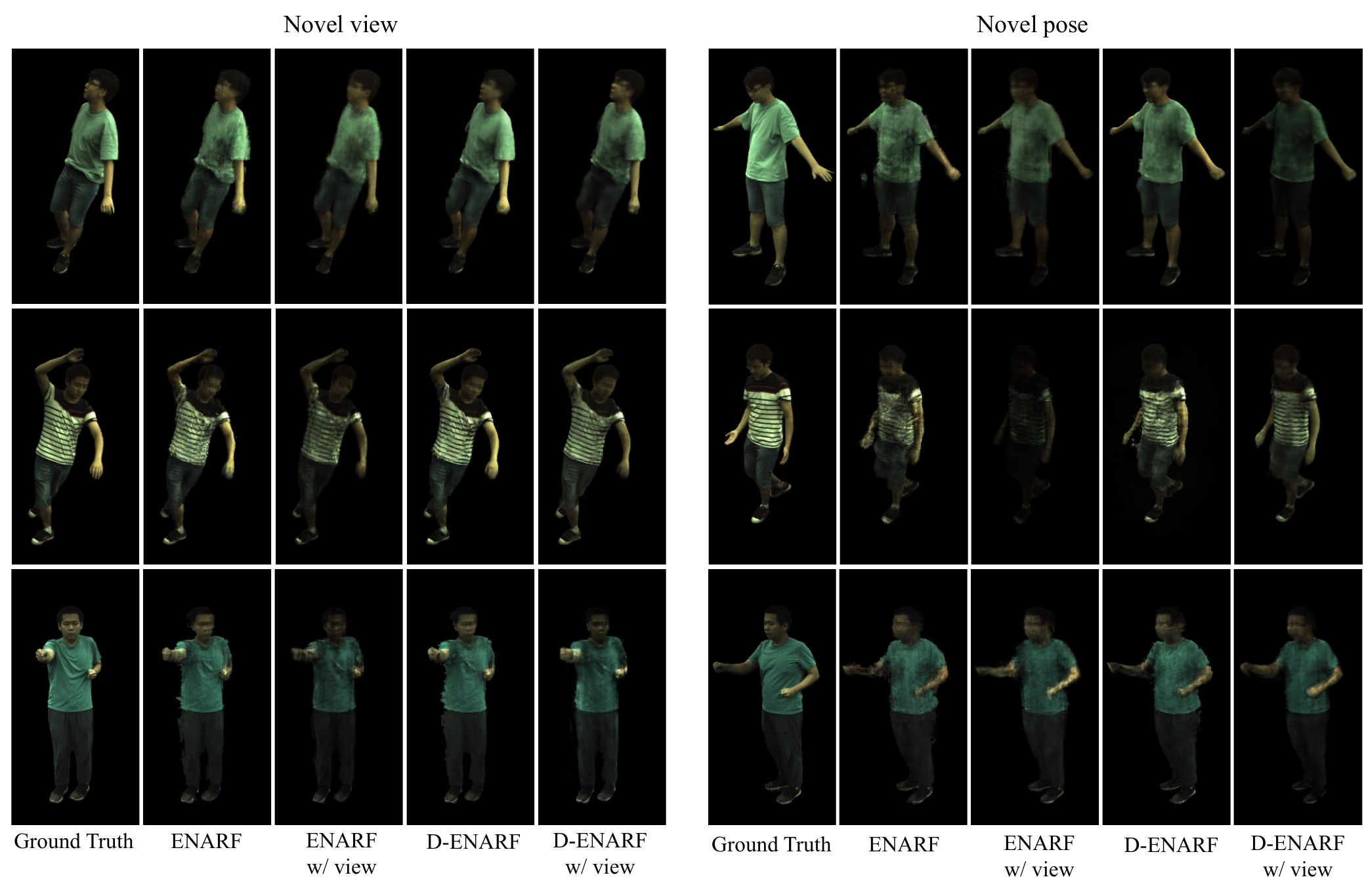}
    \caption{Ablation study on view direction.}
    \label{fig:view_direction}
  \end{figure}

  Following Efficient NeRF~\cite{chan2021efficient}, our model does not take the view direction as input, i.e. the color is not view dependent. We note that this implementation is inconsistent with the original NeRF~\cite{mildenhall2020nerf} model that takes the view direction as an input. Here, we do an ablation study on the view dependent input in our model. Specifically, the positional encoding is added to the view direction $\gamma({\bf d})$ and additionally used as the input of $G_\text{dec}$. Experimental results indicate that additional view direction input leads to darker images and degrades the performance. We thus do not use view direction as input of our models by default.
  Quantitative and qualitative comparisons are provided in Table~\ref{tab:dso_quant_ablation} and Figure~\ref{fig:view_direction}, respectively.
  
  \begin{table}[t]
    \caption{Quantitative comparison on dynamic scenes.}
    \label{tab:dso_quant_ablation}
    \centering
    \scalebox{0.8}[0.8]{
    \begin{tabular}{l|ccc|ccc}
      \hline
        & \multicolumn{3}{|c|}{Novel view} & \multicolumn{3}{|c}{Novel pose} \\
        & PSNR$\uparrow$ & SSIM$\uparrow$ & LPIPS$\downarrow$ &PSNR$\uparrow$ & SSIM$\uparrow$ & LPIPS$\downarrow$ \\ \hline \hline
        ENARF&\underline{31.94}&0.9655&0.04792&29.66&\underline{0.953}&0.05702\\
        D-ENARF&{\bf 32.93}&{\bf 0.9713}&{\bf 0.03718}&{\bf 30.06}&0.9396&\underline{0.05205}\\ \hline
        ENARF w/ view&29.98&0.9603&0.05129&28.42&0.9497&0.06005\\
        D-ENARF w/ view&31.05&\underline{0.9668}&\underline{0.04162}&29.3&{\bf 0.9563}&{\bf 0.04711}\\
      \hline
    \end{tabular}}
  \end{table}
  
  \section{Comparison with NeuralActor}
\begin{figure}[t]
    \centering
    \includegraphics[width=0.9\linewidth]{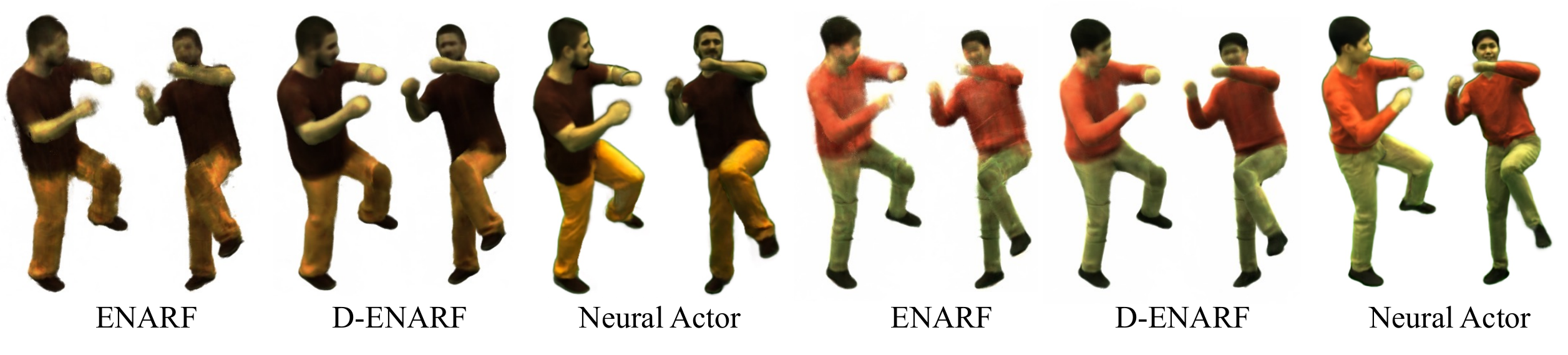}
    \caption{
Qualitative results in novel pose synthesis, compared between ENARF, D-ENARF, and NeuralActor~\cite{liu2021neural}.
The results of NeuralActor are drawn directly from Figure~4 of the original NeuralActor paper~\cite{liu2021neural}. The results of ENARF and D-ENARF are generated using similar poses and views as in NeuralActor. 
    }
    \label{fig:na_comparison}
  \end{figure}

  In this section, we compare ENARF with another state-of-the-art human image synthesis model NeuralActor~\cite{liu2021neural}. Since the training code of NeuralActor is not publically available, we trained ENARF and D-ENARF on two sequences (S1, S2) of the DeepCap dataset~\cite{marc2020deepcap} following NeuralActor. We then compared them with the corresponding results reported in the NeuralActor paper. NeuralActor uses richer supervision, such as the ground truth UV texture map of SMPL mesh for each frame. The qualitative results on novel pose synthesis are shown in Figure~\ref{fig:na_comparison}. 
  Without deformation modeling, ENARF tends to produce jaggy images and performs the worst due to the enormous non-rigid deformation in training frames. D-ENARF can alleviate the problem by learning the deformation field and can produce plausible results. Compared to NeuralActor, D-ENARF does not generate fine details such as wrinkles in clothing or facial expressions. Still, this gap is acceptable, given that NeuralActor uses GT SMPL meshes and textures for training.   
  We cannot reproduce the quantitative evaluation of NeuralActor~\cite{liu2021neural} (in Table~2) because some implementation details, such as the foreground cropping method, are not publicly available. We thus skip the quantitative comparison with NeuralActor.
 
  \section{Implementation Details of ENAFR-GAN}
  
  \subsection{Bone Region Loss $\mathcal L_\text{bone}$ (Section 3.5)}
  
\begin{figure}[t]
    \centering
    \includegraphics[width=0.6\linewidth]{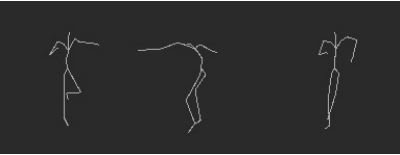}
    \caption{Examples of bone images.}
  \label{fig:bone_img}
  \end{figure}

  One obvious positional constraint for the foreground object is that it should be generated to cover at least the regions of bones defined by the input pose configuration. This motivates us to propose a bone region loss $\mathcal L_\text{bone}$ on the foreground mask $M$ to facilitate model training.
  First, we create a skeletal image $B$ from an input pose configuration. Examples of $B$ are visualized in Figure~\ref{fig:bone_img}.
  The skeletal image $B$ is an image in which each joint and its parent joint are linked by a straight line of 1-pixel width. 
  The bone region loss $\mathcal L_\text{bone}$ is then defined to penalize any overlaps between the background region ($1 - M$) and the bone regions. 
  \begin{equation}
    \mathcal L_{\text{bone}} = \frac{\sum (1 - M)^2B}{\sum B}.
  \end{equation}
  
\begin{figure}[t]
    \centering
    \includegraphics[width=0.8\linewidth]{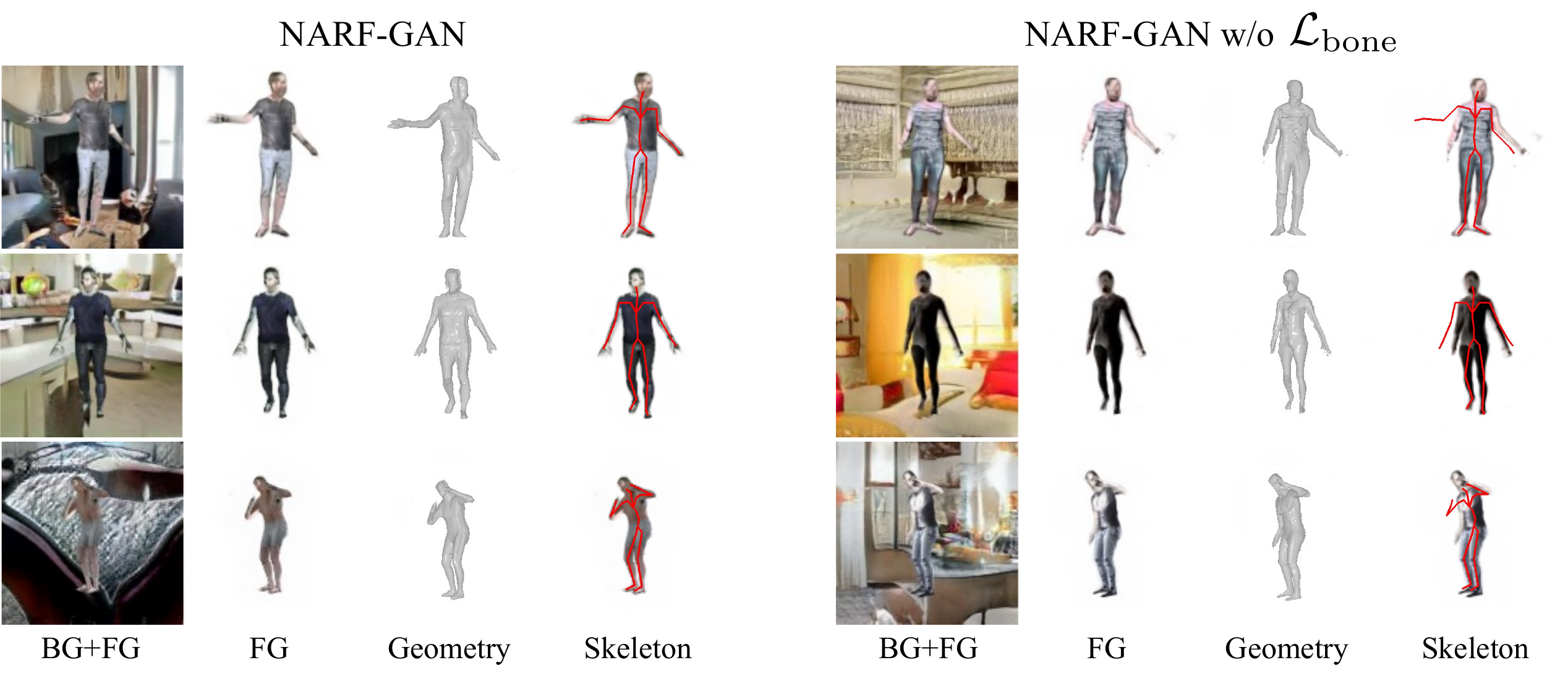}
    \caption{Ablation study on $\mathcal L_\text{bone}$. From left to right, generated images, foreground images, geometry, and foreground images with input skeletons are visualized.}
  \label{fig:l_bone_ablation}
  \end{figure}

  We show the comparison results of training with or without $\mathcal L_\text{bone}$ in Table~\ref{tab:gan_quant_appendix} and Figure~\ref{fig:l_bone_ablation}.
  Although the foreground image quality is comparable, Figure~\ref{fig:l_bone_ablation} shows that the generated images are not well aligned with the input pose without $\mathcal L_\text{bone}$.
  In Table~\ref{tab:gan_quant_appendix}, we can see that PCKh@0.5 metric becomes worse without $\mathcal L_\text{bone}$.
  
  \begin{table}[t]
    \caption{Quantitative comparison on generative models. * indicates that the methods are modified from the cited papers.}
    \label{tab:gan_quant_appendix}
    \centering
    \scalebox{0.8}[0.8]{
    \begin{tabular}{l|cccc}
      \hline
        & FID$\downarrow$ & FG-FID$\downarrow$ & Depth$\downarrow$ & PCKh@0.5$\uparrow$ \\ \hline \hline
  ENARF-GAN&22.6&21.3&8.8&0.947\\ 
  ENARF-GAN w/o $\mathcal L_\text{bone}$&24.3&21.0&14.8&0.863\\ 
      \hline
    \end{tabular}}
  \end{table}

  \subsection{Training Details of ENARF-GAN}
  We set the dimension of ${\bf z}_\text{tri}$ and ${\bf z}_b$ to 512, and ${\bf z}_\text{ENARF}$ to 256.
  We use the Adam optimizer with an equalized learning rate~\cite{karras2017progressive} of 0.0004. We set the batch size to 12 and train the model for 300,000 iterations. The training takes 4 days on a single A100 GPU.
  In the testing phase, a 128$\times$128 image is rendered in 80ms (about 12 fps) using a single A100 GPU.
  
  \subsection{Shifting Regularization}
  To prevent the background generator from synthesizing the foreground object, we apply shifting regularization~\cite{bielski2019emergence} for the background generator.
  We randomly shift and crop the background images before overlaying foreground images on them.
  If the background generator synthesizes the foreground object, the shifted images can be easily detected by the discriminator $D$,
  which encourages the background generator to focus only on the background.
  We generate $128 \times 256$ background images and randomly crop $128 \times 128$ images.
  
  \section{Pose Consistency Metric}
  We follow the evaluation metric in the contemporary work GNARF~\cite{bergman2022generative} to evaluate the consistency between the input pose and the pose of the generated image. Specifically, we use an off-the-shelf 2D keypoint detector~\cite{mmpose2020} pre-trained on the MPII human pose dataset~\cite{andriluka20142d} to detect 2D keypoints in both generated and real images with the same poses and compare the detected keypoints under the metric of PCKh@0.5~\cite{andriluka20142d}. We discard keypoints with low detection confidence ($<0.8$) and only compare keypoints that are confident in both generated and real images.
  
  \section{Truncation Trick (Section 4.2)}
\begin{figure}[t]
    \centering
    \includegraphics[width=0.9\linewidth]{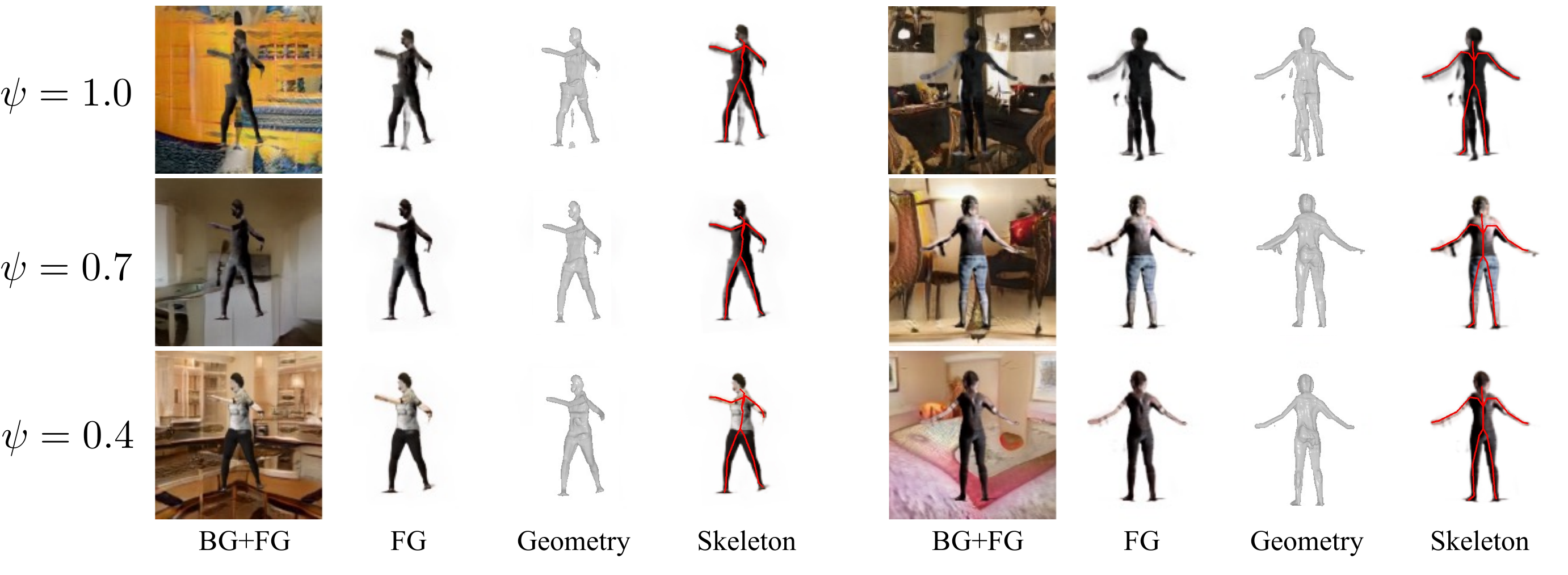}
    \caption{Truncation trick.}
  \label{fig:truncation}
  \end{figure}

  The truncation trick~\cite{karras2019style} can improve the quality of the images by limiting the diversity of the generated images.
  Figure~\ref{fig:truncation} shows the results of generating images with the truncation $\psi$ for the tri-plane generator $G_\text{tri}$ set to 1.0, 0.7, and 0.4.
  When truncation $\psi$ is set to 1.0, multiple legs and arms will be generated. Smaller $\psi$ helps generate more plausible appearance and shapes of the object.
  
  \section{StyleNARF (Section 4.2)}
  StyleNARF is a combination of NARF~\cite{noguchi2021neural} and StyleNeRF~\cite{gu2022stylenerf}.
  To reduce the computational complexity, it first generates low-resolution features with NARF and then upsamples the features to a higher resolution with a CNN-based generator.
  Similar to ENARF, StyleNARF can only generate foreground objects, so a StyleGAN2 based background generator $G_\text{b}$ is used as in ENARF-GAN.
  First, we sample latent vectors from a normal distribution, ${\bf z} = ({\bf z}_\text{NARF}, {\bf z}_\text{b}, {\bf z}_\text{up}) \sim \mathcal N (0, I)$, where ${\bf z}_\text{NARF}$ is a latent vector for NARF, ${\bf z}_\text{b}$ is a latent vector for background, and ${\bf z}_\text{up}$ is a latent vector for the upsampler. Then the NARF model $G_\text{NARF}$ generates low-resolution foreground feature ${\bf F}_f$ and mask ${\bf M}_f$, and $G_\text{b}$ generates background feature ${\bf F}_b$.
  \begin{equation}
    {\bf F}_f, {\bf M}_f = G_\text{NARF}({\bf z}_\text{ENARF}, \{l_k, R_k, {\bf t}_k\}_{k=1:K}), {\bf F}_b = G_\text{b}({\bf z}_b).
  \end{equation}
  The foreground and background are combined using the generated foreground mask ${\bf M}_f$ at low resolution.
  \begin{equation}
    {\bf F} = {\bf F}_f + {\bf F}_b * (1 - {\bf M}_f).
  \end{equation}
  The upsampler $G_\text{up}$ upsamples the feature ${\bf F}$ based on the latent vector ${\bf z}_\text{up}$ and generates the final output ${\bf C}$.
  \begin{equation}
    {\bf C} = G_\text{up}({\bf F}, {\bf z}_\text{up}).
  \end{equation}
  All layers are implemented with Modulated Convolution~\cite{karras2020analyzing}.
  
  \section{Datasets}
  We use images at resolution $128 \times 128$ for GAN training.
  
  \subsection{SURREAL}
  We crop the first frame of all videos to $180\times 180$ and resize them to $128\times 128$ so that the pelvis joint is centered. 68033 images are obtained.
  
  \subsection{AIST++}
  The images are cropped to $600 \times 600$ so that the pelvis joint is centered and then resized to $128 \times 128$.
  We sample 3000 frames for each subject, resulting in 90000 images in total.
  
  \subsection{MSCOCO}
  First, we select the persons whose entire body is almost visible according to the 2D keypoint annotations in MSCOCO. Each selected person is cropped by a square rectangle that tightly encloses the person and it is resized to $128 \times 128$. The number of collected samples is 38727.
  
  \section{Pose Distribution}
  Two pose distributions are used in our experiments. One is the CMU pose distribution, which consists of 390k poses
  collected in the CMU Panoptic dataset.
  We follow the pose pre-processing steps in the 
  SURREAL dataset, where the distance between the person and the camera is randomly distributed in a normal distribution of mean 7 meters and variance 1 meter. The pose rotation around the z-axis is uniformly distributed between 0 and $2\pi$.
  
  Another pose distribution used in our experiments is a random pose distribution. First, a multivariate normal distribution is fitted to the angle distribution of each joint under the 390k pose samples
  of the CMU Panoptic dataset. Then, poses are randomly sampled from the learned multivariate normal distribution
  and randomly rotated so that the pelvis joint is directly above the medial points of the right and left plantar feet.
  
\end{document}